\documentclass{article}

\PassOptionsToPackage{numbers, compress}{natbib}
    \usepackage[preprint]{neurips_2020}

\usepackage[utf8]{inputenc} % allow utf-8 input
\usepackage[T1]{fontenc}    % use 8-bit T1 fonts
\usepackage{hyperref}       % hyperlinks
\usepackage{url}            % simple URL typesetting
\usepackage{booktabs}       % professional-quality tables
\usepackage{amsfonts}       % blackboard math symbols
\usepackage{nicefrac}       % compact symbols for 1/2, etc.
\usepackage{microtype}      % microtypography

\usepackage{float}
\usepackage{tikz}
\usepackage{amsmath}
\usepackage{enumitem}
\usepackage{wrapfig}
\DeclareMathOperator{\E}{\mathbb{E}}
\newtheorem{definition}{Definition}[section]

\usepackage{algorithm}
\usepackage[noend]{algpseudocode}

\title{Improving Fair Predictions Using Variational Inference In Causal Models}

\author{
   Rik ~Helwegen\thanks{Part of this work was done at Statistics Netherlands (CBS).}\\
   University of Amterdam\\
   \texttt{helwegen.rc@gmail.com} \\
  \And
    Christos ~Louizos\\
  University of Amterdam\\
  \texttt{c.louizos@uva.com} \\
  \And
  Patrick ~Forr\'e\\
  University of Amsterdam\\
    \texttt{p.d.forre@uva.nl} \\
}

\begin{document}

\maketitle

\begin{abstract}
The importance of algorithmic fairness grows with the increasing impact machine learning has on people's lives.
Recent work on fairness metrics shows the need for causal reasoning in fairness constraints.
In this work, a practical method named \textit{FairTrade} is proposed for creating flexible prediction models which integrate fairness constraints on sensitive causal paths.
The method uses recent advances in variational inference in order to account for unobserved confounders.
Further, a method outline is proposed which uses the causal mechanism estimates to audit black box models.
Experiments are conducted on simulated data and on a real dataset in the context of detecting unlawful social welfare.
This research aims to contribute to machine learning techniques which honour our ethical and legal boundaries. 
\end{abstract}

\section{Introduction}
\label{introduction}
Over the past decade, the performance increase of machine learning (ML) has stimulated applications within various fields \cite{lecun2015deep}, including the social domain.
Of particular interest for this work are applications in which the data points consist of people's profiles, and the model outcomes have direct consequences for the individuals these profiles belong to.
ML applications with high individual impact include loan approval \cite{mahoney2007method}, police screenings \cite{magee2008neighbourhood} and fraud detection.
Criticism on predictive models, such as the report by \citet{angwin2016machine}, has been increasing, and is often based on possible unwanted discrimination.
ML models differentiate between input profiles in order to make valuable classifications or predictions.
The question is which information is deemed fair and unfair to base the differentiation on, and how to effectively exclude unfair information.
A central topic within the domain of algorithmic fairness is defining the desired relation between \textit{sensitive attributes}, such as gender and ethnicity, and model outcomes \cite{loftus2018causal}.
Metrics based on observational and predictive distributions structurally showed difficulties in capturing fairness properly \cite{dedeo2014wrong}, motivating the introduction of the causality-based metric \textit{counterfactual fairness} \cite{kusner2017counterfactual}.
This metric captures the intuition that a change in value of the sensitive attribute, including the causal results this would have for other attributes, should have no effect on the model outcomes.
Recently this definition is further refined in order to deal with situations in which the outcome depends on the sensitive variable through both fair and unfair causal relations \cite{chiappa2018path, nabi2018fair}.
We make the following three contributions:
\begin{enumerate}[topsep=0pt,itemsep=-0.5ex,partopsep=1ex,parsep=1ex]
    \item We propose the FairTrade method, which improves prediction models with fairness constraints on causal pathways by addressing unobserved confounders of the covariates.
    \item Insight in the applicability of novel techniques is created through a large scale real data experiment on detecting risk profiles for unlawful social welfare.
    \item A method is proposed to use the estimates of causal mechanisms to audit black box models. This way, existing (non-causal) models can be scored on causality-based fairness metrics.
\end{enumerate}
This work continues by reviewing relevant literature in section \ref{background}.
In section \ref{fairtrade_method}, we introduce our method to incorporate causality-based fairness constraints in predictive models. 
Subsequently, we validate and explore the method using synthetic and real data in section \ref{experiments}.
Section \ref{black-box-scoring} provides the outline for auditing black box models, given a situation in which an accurate estimate of causal mechanisms can be obtained.
Finally, a discussion on the method follows, after which we end by putting the research in context of its broader impact.

\section{Background}
\label{background}

\subsection{Causality}
In causality theory the goal is to study causal relations rather than relying on statistical correlation only.
The use of this can be observed when analysing the results of intervening in a system.
Consider an \textit{intervention} as an external force, setting a variable to a particular value.
Some variables might be causally influenced by this interventions, while other variables remain unaltered. 
This behaviour can be formalised using theory introduced by \citet{pearl2009causality}.
Causal relations are visually expressed in a \textit{graph}. 
This work is bound to using Directed Acyclic Graphs (DAG), which excludes cyclic relations, but may include explicitly modelled unobserved variables.
Bold symbols from here on indicate a set, e.g. $\boldsymbol{X}=(X_1,..,X_d)$. 
The edge $X \rightarrow Y$ of $\mathcal{G}$ reflects the direct effect of node $X$ on node $Y$, such that $X$ becomes a parent of $Y$, and the two nodes are considered adjacent.
A path is a tuple of nodes which are all unique and successively adjacent. 
If the directed relations do not collide in the path, it is considered a directed path.

For the link between graphs and causal modelling, we follow the definition by \citet{peters2017elements} of \textit{Structural Causal Models} (SCM).
The SCM can be used to reason out possible outcomes after an intervention.
Further, if we have already observed an outcome and reason about what \textit{would have happened} after an intervention, we speak about counterfactuals.
\citet{pearl2009causality} introduces \textit{do-Calculus}, in order to describe such distributions. 
Interventional and counterfactual distributions are expressed by using functional notation for the outcome variables, in which the input reflects possible interventions.
Capital letters describe random variables, and small letters are used for value realisations.
The random variable $Y$, under intervention of $A=a'$ will be denoted as $Y(a')$.
Writing $Y(A)$, using capital $A$, indicates the random variable $Y$ when intervening $A$ on its naturally obtained value, which is equal to $Y$ without intervention on $A$ \cite{robins1986new}.

For a refined denotation of causal effects we consider \textit{Path specific effects} (PSE) \cite{pearl2001direct, avin2005identifiability}.
The effect of the directed path $\pi$, going from $X$ to $Y$, is defined as the change in $Y$ when changing $X$ from base value $x'$ to $x$ along $\pi$.
More formally, we rely on the inductive rule by \citet{shpitser2013counterfactual} to obtain the correct counterfactual, see def. \ref{def pse}.
For example, considering Figure \ref{fig:fairtrade_graph} (c), the estimate of the PSE of $A \rightarrow R \rightarrow Y$ when intervening $A=a$, with $A=a'$ as reference, would be; $\E[Y(a', Z, B, X(a'), R(a, X(a')))] - \E[Y(a', Z, B, X(a'), R(a'))]$. 
\begin{definition}{(Path Specific Effect (PSE))}\\
    \label{def pse}
    The PSE for a set of directed paths $\boldsymbol{\pi}$ from $X$ to $Y$ when changing $X$ from $x'$ to $x$ is the expected change in the nested counterfactual value of $Y$ \cite{shpitser2013counterfactual}. 
    Along paths in $\boldsymbol{\pi}$ the effect of $X$ is propagated as the active value $x$, whereas all other paths propagate its effect as base value $x'$.
\end{definition}

\subsection{Variational Inference}
As large data profiles and unobserved confounders often make exact inference of the causal relations impossible or intractable, we make use of variational inference. 
This way, a replacement for the unobserved confounders is inferred per data point, and subsequently used in the reconstruction model.
Instead of optimising the likelihood of observing the data under the learned model, the objective becomes to maximise a (variational) \textit{lower bound} of this likelihood.
\citet{kingma2013auto} and \citet{rezende2014stochastic} introduced the technique known as the variational autoencoder (VAE). 
An encoder for obtaining the latent space and a decoder, reconstructing the data, are simultaneously optimised.
This \textit{inference} model and \textit{generative} model can be parameterised as neural networks.
The reparameterization trick \cite{kingma2013auto} allows the gradient to flow back through the sample of the inferred latent distributions.
The model is evaluated using the log likelihood of observing the data under the reconstructed distributions. 
The VAE can recover a large variety of latent distributions, as shown by \citet{tran2015variational}, and repeatedly shows successful applications in various domains \citep{rezende2016unsupervised, chung2015recurrent, gregor2015draw}.

We incorporate the work by \citet{louizos2017causal}, introducing the causal effect variational autoencoder (CEVAE), for the estimation of causal effects.
The latent space represents the unobserved confounders, which are latent variables affecting multiple covariates in the data profiles.
The generative network, or decoder, represents the causal data generative process, and is structured in accordance with the assumed causal graph.
These deep latent-variable models create an efficient way of estimating the relations in a causal graph, making it possible to take on real data scenario's with unobserved confounders and a large number of covariates. 
Compared to exact causal inference, a much broader range of settings becomes feasible to handle given restricted computational power.
For our goal of improving on fairness in practical scenarios, these benefits can outweigh the fact that the CEVAE is an approximation with limited guarantees of recovering the true underlying causal relations. 
A case specific analysis of the causal structure, optimisation performance and feasible alternatives is important while considering this trade-off.

\subsection{Fairness}
The question of `What it means for algorithmic decisions to be fair?' requires a discussion involving different disciplines.
In this work, the focus is on preventing a statistical model to discriminate on the basis of sensitive information.
This work does not discuss related problems such as difference in model uncertainty for different groups, or preventing feedback loops when introducing an algorithm into an existing process.
Which causal paths should be marked unfair is also out of scope, although the proposed methods are meant to support such decision making by allowing comparison between outcomes under different fairness restrictions.

In this work we hold on to the convention of $A$ being the sensitive attribute and $Y$ the outcome variable.
One way of pursuing fairness is to leave out the sensitive variable from the input of the predictive model.
As proxy variables can propagate sensitive information, this method can be ineffective and even lead to less fair policies.
Several \textit{observational} fairness metrics are proposed in previous literature, such as \textit{Statistical Parity}: constraining $p(\hat{Y}=y|A=a) = p(\hat{Y}=y|A=a')$ for all $a, y$, and \textit{Equalised Odds} with the condition $p(\hat{Y}=y|A=a,Y=y) = p(\hat{Y}=y|A=a',Y=y)$ for all $a, y$.
As explained by \citet{dedeo2014wrong}, \citet{berk2018fairness} and \citet{kusner2017counterfactual} individual observational metrics often lack to comply with our intuition of fairness in a variety of situations, and metrics are often not compatible with the exception of using trivial solutions such as random guessing.
This motivates the need for fairness metrics in which the causal relations between variables are considered.
Counterfactual Fairness, as proposed by \citet{kusner2017counterfactual}, forms the basis of a causality based fairness metrics.
\begin{definition}{(Counterfactual Fairness)}\\
    The predictive distribution $\hat{Y}$ is counterfactually fair if
    \begin{align}
        \begin{gathered}
        p(\hat{Y}(a,x)=y|X=x, A=a) = p(\hat{Y}(a',x)=y|X=x, A=a)\quad \forall y, x, a, a'
        \end{gathered}
    \end{align}
\end{definition}
\citet{nabi2018fair} and \citet{chiappa2018path} build on this metric to define fairness in case the sensitive attribute is related to the outcomes through fair and unfair paths.
\textit{Path specific counterfactual fairness} forms the basis for fairness in the proposed method.
\begin{definition}{(Path Specific Counterfactual Fairness)}\\
    \label{def_pscf}
    The predictive distribution $\hat{Y}$ is path specific counterfactual fair with respect to the set of paths $\boldsymbol{\pi}$ if the counterfactual predictive distribution is not affected by the PSE for the set of paths $\boldsymbol{\pi}$ (def \ref{def pse}). 
\end{definition}
The intuition behind this metric is that, given a causal graph, information of the sensitive variable cannot influence the outcome distribution via unfair paths. 
\section{FairTrade method}
\label{fairtrade_method}
This section provides an overview of the general steps and considerations of the method. 
The implementation details of the experiments, described in the following section and appendix, are suitable as practical examples. 
The FairTrade method follows three consecutive steps:
\begin{align*}
    \text{1) \textbf{Model} causal structure} \rightarrow \quad
    \text{2) \textbf{Learn} causal mechanisms} \rightarrow \quad
    \text{3) \textbf{Train} fair prediction model}
\end{align*}
In the first step, causal and distributional assumptions are set by modelling the causal structure of the data in the form of a graph.
In the second step, the relations in the graph are approximately estimated using the CEVAE method.
Finally, the estimates of the causal relations are used selectively in order to create an auxiliary prediction model in line with the fairness requirements.
To prevent the fairness constraints from affecting the estimates of the (non-constrained) effect estimates, the steps are taken in consecutive order.
The FairTrade method requires the following assumptions to hold: 
(I) a correct causal structure, 
(II) a sufficiently good representation of all unobserved confounders,
and (III) the identifiability of unfair causal paths.

\subsection{Model causal structure}
\begin{wrapfigure}{r}{0.7\textwidth}
        \centering
        \resizebox{235pt}{70pt}{
        \input{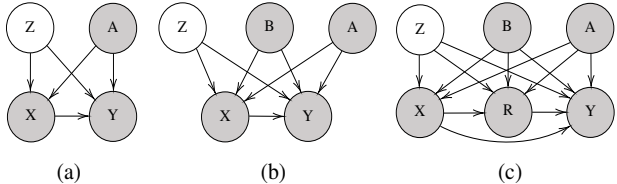}
        }
        \caption{Grey nodes are observed, white nodes unobserved. (a) The \textit{sensitive variable} (A) and the \textit{latent confounders} (Z) are parents of the \textit{covariates} (X) and the \textit{objective} (Y). In (b), \textit{base variables} (B) are added, which are independent of (A). In (c) the \textit{resolving variables} (R) are added.} 
        \label{fig:fairtrade_graph}
\end{wrapfigure}
The causal graph represents the assumed relations in the data, and is of fundamental importance to the method. 
The graph structure is case specific, and primarily based on domain knowledge.
In this step, the data is only used to confirm the implied independence relations.
The graph is restricted to be a DAG, which can include explicitly modelled unobserved confounders.
The causal structure should always be constructed as objective as possible, averting any political interference in this step of the method.
A diverse team, or input from people with different perspectives, can contribute to a broader reflection of values in the obtained domain knowledge.
Three causal structures are proposed which are applicable to many scenarios of interest, shown in Figure \ref{fig:fairtrade_graph}. 
The provided structures treat the sensitive attribute as root variable. 
Variables like ethnicity and gender are indeed often at the base of a causal structure.
A factor like \textit{mother's ethnicity}, which influences ethnicity, is part of the sensitive information, and should be included in the set of sensitive attributes \cite{kusner2017counterfactual}.
In other cases there might be factors of influence on the sensitive attribute which are not sensitive themselves. 
For example, consider prohibiting \textit{prior convictions} from influencing \textit{hiring}.
This requires case specific adjustments of the method.

The first model (Fig. \ref{fig:fairtrade_graph}a) assumes that all covariates are influenced by the sensitive attribute.
This is the most conservative setup in the sense that we might be inefficient by including redundant relations, but we will not create bias by omitting existing effects coming from the sensitive variable
\cite{heij2004econometric}.
As it is unrealistic to assume all variables in the complex generation of data profiles are observed, the confounder Z takes up the role of unobserved background variable.
The latent space captures all background information apart from sensitive information, which is covered by $A$ itself.
Note that this variable is a confounding factor for the covariates and the outcomes, but not for the sensitive treatment $A$. 
This non-sensitive covariate confounder thus differs from the confounder as defined by \citet{vanderweele2013definition}.
The graphs (b) and (c) in Figure \ref{fig:fairtrade_graph} are variations of the first setup. 
In the case of (b) a node for \textit{base variables} is added, addressing variables which are known to be independent of A, but which do affect other variables. 
In (c) the node R is introduced for \textit{resolving variables}.
The distinction between resolving variables and other covariates is made in order to allow for fair path specific effects which include descendants of $A$.
As resolving variables are often closely related to the outcome, the node is positioned in between $X$ and $Y$. 
As an example, historical observations of the outcome itself could be resolving variables.

\subsection{Learn causal mechanisms}
In the second step of the FairTrade method, the observed data and assumptions are combined in order to approximately learn the causal mechanisms, which is done using the CEVAE technique proposed by \citet{louizos2017causal}. 
As most applications of interest include many covariates and unobserved confounders, this approximate but tractable method is preferred over exact causal inference methods.
Random variables are modelled as probability distributions, and the relations between variables are parameterised as neural networks, hereby allowing for flexible relations. 
The objective Y is not included as input to the inference network in order to allow for out-of-sample prediction. 
The generative network becomes a chain of networks structured according to the causal graph.
The CEVAE is optimised using a variational lower bound, following the lines of \citet{kingma2013auto} and \citet{louizos2017causal}. 
The lower bound for the graph in Figure \ref{fig:fairtrade_graph} (c) is defined $\mathcal{L}_{1c} = \mathcal{L}_{Reg} + \mathcal{L}_{Rec}$ with the regularisation and reconstruction terms shown in equation \ref{reg_lower} and \ref{rec_lower} respectively.
\begin{align}
    \label{reg_lower}
    \mathcal{L}_{Reg} &= \sum_{i=1}^N\E_{q(z_i|a_i,b_i,x_i,r_i)}[\log p(z_i)   - \log q(z_i|a_i,b_i,x_i,r_i)]\\
    \label{rec_lower}
    \mathcal{L}_{Rec} &= \sum_{i=1}^N\mathbb{E}_{q(z_i|...)}[\log p(x_i|b_i,a_i,z_i) 
    + \log p(r_i|b_i,a_i,x_i,z_i)+ \log p(y_i|b_i,a_i,r_i,x_i,z_i)]
\end{align}
Where $i$ indicates the observation.
A SGD-based optimiser, such as ADAM \cite{kingma2014adam}, is used to optimise the CEVAE model.
To estimate the sensitive (treatment) effect in the generative model, following \citet{louizos2017causal}, the networks obtain a TAR structure \cite{shalit2017estimating} with separate heads for the values of the sensitive variable.
Excluding unfair paths requires their identifiability.
PSE may not be identified, even in the absence of unobserved confounders \cite{shpitser2013counterfactual}.
We consider the cases in which a collection of all paths from $A$ to $Y$ is marked as unfair, and we use the recanting witnesses criterion \cite{avin2005identifiability} for deriving identifiability, in line with \citet{zhang2016causal}.
For Figure 1c, we obtain the result that all possible selections of unfair paths are identifiable, with one type of exception. 
The paths $A \rightarrow X \rightarrow Y$ and $A \rightarrow X \rightarrow R \rightarrow Y$ are only identifiable `together'; sets of paths containing one of the two are not identifiable with $X$ acting as recanting witness. 
This implies that if one of these paths is deemed unfair and the other is not, the identifiability assumption does not hold, and the FairTrade method is not applicable in its standard form.

\subsection{Train fair prediction model}
The trained CEVAE is an estimate of causal relations in the observed data. 
The next step is to make a predictive model of Y under the posed fairness restrictions.
This model is created in line with \citet{kusner2017counterfactual} and \citet{loftus2018causal}, by training an auxiliary model for which the input selection determines the imposed fairness constraints.
The input of the auxiliary model cannot include information from unfair causal paths. 
It can include non-descendants of $A$ and descendants of $A$ which are not part of unfair paths. 
If descendants of $A$ are part of both fair and unfair paths, a nested counterfactual can include inputs from both the observed and altered value of $A$.
For example, if in the graph of Figure \ref{fig:fairtrade_graph} (c) the path $A\rightarrow R \rightarrow Y$ would be deemed fair (resolved) but none of the other paths from $A$ to $Y$ are, one can use $\{Z, B, R(X(a'),Z,B,A)\}$ with baseline value $a'$ as input for the auxiliary model. 
This way, the fair part of $R$ is used and the path $A\rightarrow X\rightarrow R \rightarrow Y$ is `blocked' through (nested) intervention on $A$ propagated via $X$.

\subsection{Related work}
The FairTrade method uses a causal latent variable approach, similar to approaches by \citet{loftus2018causal} and \citet{kusner2017counterfactual}.
Unlike these approaches, the FairTrade method makes the link with the CEVAE \cite{louizos2017causal} to address unobserved confounders. 
\citet{madras2018learning} likewise do so, but do not proceed to counterfactual fairness or path specific counterfactual fairness.
\citet{kilbertus2017avoiding} introduce resolving variables in their work, but take on a different approach in resolving effects, being less refined compared to considering path specific effects.
The work by \citet{nabi2018fair} requires calculation of the path specific effects, which is often not feasible in practical applications. 
Further, like \citet{nabi2019optimal}, this work performs a likelihood optimisation rather than optimising a variational lowerbound.
\citet{chiappa2018path} propose a related method to obtain path specific counterfactual fairness.
One difference is that this work uses latent spaces per observed variable, and is thus not applicable with unobserved confounders in its proposed form.
It might be possible to extend the method in this regard.
Other differences include the forced independence of the latent spaces with the sensitive variable, using a MMD \cite{gretton2012kernel} term, and the simultaneous optimisation of the causal estimate and predictive network.
We refrain from these approaches in order to minimise interference with the causal inference, which for example might `work around' fairness restrictions by overestimating non restricted paths.
One approach to deal with unobserved confounders in causal inference is recently proposed by \citet{wang2019blessings} called the \textit{deconfounder}, which raised several comments \cite{ogburn2019comment, d2019comment}.
Like in the FairTrade method, the deconfounder considers the recovery of unobserved confounders.
\section{Experiments}
\label{experiments}
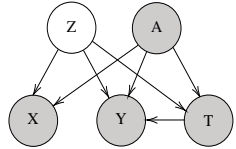
\begin{wrapfigure}{r}{0.4\textwidth}
    \centering
    \resizebox{90pt}{55pt}{
    \tikzset{every picture/.style={line width=0.75pt}} %set default line width to 0.75pt        

\begin{tikzpicture}[x=0.75pt,y=0.75pt,yscale=-1,xscale=1]

%Shape: Circle [id:dp8882202064361273] 
\draw   (353,35) .. controls (353,21.19) and (364.19,10) .. (378,10) .. controls (391.81,10) and (403,21.19) .. (403,35) .. controls (403,48.81) and (391.81,60) .. (378,60) .. controls (364.19,60) and (353,48.81) .. (353,35) -- cycle ;
%Shape: Circle [id:dp3972462279050821] 
\draw  [fill={rgb, 255:red, 207; green, 204; blue, 204 }  ,fill opacity=1 ] (440.5,35.5) .. controls (440.5,21.69) and (451.69,10.5) .. (465.5,10.5) .. controls (479.31,10.5) and (490.5,21.69) .. (490.5,35.5) .. controls (490.5,49.31) and (479.31,60.5) .. (465.5,60.5) .. controls (451.69,60.5) and (440.5,49.31) .. (440.5,35.5) -- cycle ;
%Straight Lines [id:da39202659843045606] 
\draw    (364.3,56.2) -- (339.3,99.47) ;
\draw [shift={(338.3,101.2)}, rotate = 300.02] [color={rgb, 255:red, 0; green, 0; blue, 0 }  ][line width=0.75]    (10.93,-3.29) .. controls (6.95,-1.4) and (3.31,-0.3) .. (0,0) .. controls (3.31,0.3) and (6.95,1.4) .. (10.93,3.29)   ;
%Straight Lines [id:da9512001522500872] 
\draw    (482.3,54.2) -- (509.31,101.46) ;
\draw [shift={(510.3,103.2)}, rotate = 240.26] [color={rgb, 255:red, 0; green, 0; blue, 0 }  ][line width=0.75]    (10.93,-3.29) .. controls (6.95,-1.4) and (3.31,-0.3) .. (0,0) .. controls (3.31,0.3) and (6.95,1.4) .. (10.93,3.29)   ;
%Straight Lines [id:da5870009072296218] 
\draw    (390.3,57.2) -- (415.32,101.46) ;
\draw [shift={(416.3,103.2)}, rotate = 240.52] [color={rgb, 255:red, 0; green, 0; blue, 0 }  ][line width=0.75]    (10.93,-3.29) .. controls (6.95,-1.4) and (3.31,-0.3) .. (0,0) .. controls (3.31,0.3) and (6.95,1.4) .. (10.93,3.29)   ;
%Straight Lines [id:da517385826313759] 
\draw    (446.3,52.2) -- (360.96,110.08) ;
\draw [shift={(359.3,111.2)}, rotate = 325.86] [color={rgb, 255:red, 0; green, 0; blue, 0 }  ][line width=0.75]    (10.93,-3.29) .. controls (6.95,-1.4) and (3.31,-0.3) .. (0,0) .. controls (3.31,0.3) and (6.95,1.4) .. (10.93,3.29)   ;
%Shape: Circle [id:dp6295729985825067] 
\draw  [fill={rgb, 255:red, 207; green, 204; blue, 204 }  ,fill opacity=1 ] (313.67,126) .. controls (313.67,112.19) and (324.86,101) .. (338.67,101) .. controls (352.47,101) and (363.67,112.19) .. (363.67,126) .. controls (363.67,139.81) and (352.47,151) .. (338.67,151) .. controls (324.86,151) and (313.67,139.81) .. (313.67,126) -- cycle ;
%Shape: Circle [id:dp10974314984293798] 
\draw  [fill={rgb, 255:red, 207; green, 204; blue, 204 }  ,fill opacity=1 ] (493.83,125.83) .. controls (493.83,112.03) and (505.03,100.83) .. (518.83,100.83) .. controls (532.64,100.83) and (543.83,112.03) .. (543.83,125.83) .. controls (543.83,139.64) and (532.64,150.83) .. (518.83,150.83) .. controls (505.03,150.83) and (493.83,139.64) .. (493.83,125.83) -- cycle ;
%Shape: Circle [id:dp5869435274255634] 
\draw  [fill={rgb, 255:red, 207; green, 204; blue, 204 }  ,fill opacity=1 ] (404,125.67) .. controls (404,111.86) and (415.19,100.67) .. (429,100.67) .. controls (442.81,100.67) and (454,111.86) .. (454,125.67) .. controls (454,139.47) and (442.81,150.67) .. (429,150.67) .. controls (415.19,150.67) and (404,139.47) .. (404,125.67) -- cycle ;
%Straight Lines [id:da5044943355726361] 
\draw    (455.3,58.2) -- (437.09,100.36) ;
\draw [shift={(436.3,102.2)}, rotate = 293.36] [color={rgb, 255:red, 0; green, 0; blue, 0 }  ][line width=0.75]    (10.93,-3.29) .. controls (6.95,-1.4) and (3.31,-0.3) .. (0,0) .. controls (3.31,0.3) and (6.95,1.4) .. (10.93,3.29)   ;
%Straight Lines [id:da9497551200045244] 
\draw    (399.3,48.2) -- (493.67,115.04) ;
\draw [shift={(495.3,116.2)}, rotate = 215.31] [color={rgb, 255:red, 0; green, 0; blue, 0 }  ][line width=0.75]    (10.93,-3.29) .. controls (6.95,-1.4) and (3.31,-0.3) .. (0,0) .. controls (3.31,0.3) and (6.95,1.4) .. (10.93,3.29)   ;
%Straight Lines [id:da3866822983292444] 
\draw    (494.3,125.2) -- (457.3,125.2) ;
\draw [shift={(455.3,125.2)}, rotate = 360] [color={rgb, 255:red, 0; green, 0; blue, 0 }  ][line width=0.75]    (10.93,-3.29) .. controls (6.95,-1.4) and (3.31,-0.3) .. (0,0) .. controls (3.31,0.3) and (6.95,1.4) .. (10.93,3.29)   ;

% Text Node
\draw (378,34) node  [font=\large] [align=left] {Z};
% Text Node
\draw (465.5,33.5) node  [font=\large] [align=left] {A};
% Text Node
\draw (338.67,124) node  [font=\large] [align=left] {X};
% Text Node
\draw (518.83,125.83) node  [font=\large] [align=left] {T};
% Text Node
\draw (429,123.67) node  [font=\large] [align=left] {Y};

\end{tikzpicture}
    }
    \caption{Graphical model for the IHDP semi-simulated experiment.} 
    \label{fig:ihdp_graph}
\end{wrapfigure}
\subsection{IHDP semi-simulated experiment}
\begin{wrapfigure}[22]{r}{0.5\textwidth}
    \centering
    \resizebox{200pt}{150pt}{
    \includegraphics{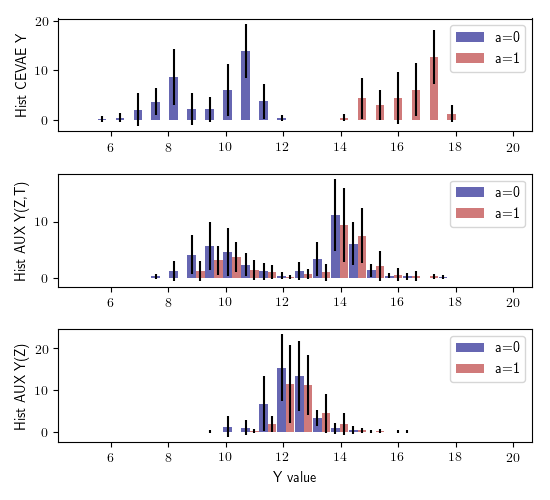}
    }
    \caption{Prediction distribution of $Y$ for models with different input criteria (std in black). The CEVAE uses $Z$, $T$ and $A$ to predict $Y$. The auxiliary model with $A$ removed from the input shows that the conditional distributions now overlap. Removing $T$ shows the two modes of the conditional distributions disappear.}
    \label{fig:pse_ihdp}
\end{wrapfigure}

In line with \citet{madras2018learning} and \citet{louizos2017causal} a semi-simulated version of the IHDP dataset \cite{hill2011bayesian} is used for analysis.
This dataset is used for analysing a medical treatment effect for low-weight premature infants, and consists of 767 observations with 25 covariates.
The covariates include information on the infants and their surroundings.
The algorithm for generating the continuous outcomes ($Y$) and the code for the experiments are provided in the appendix.
A CEVAE model is structured according to the graph in Figure \ref{fig:ihdp_graph}, staying close to \citet{louizos2017causal}.
The outcomes shown in Figure \ref{fig:pse_ihdp} are in line with the idea that effects from different paths in the model can be excluded by specifying the input of the auxiliary prediction model. 
The CEVAE model does reconstruct the modes of the true Y distribution.
Going from the first to second top graph, we see the direct effect of $A$ is excluded, making the conditional distributions overlap, but the two modes caused by the treatment ($T$) remain intact. 
Subsequently, removing the treatment from the auxiliary input merges the two modes. 

The reconstructed distribution by the CEVAE is not as fat-tailed as the true data distribution of Y, which might indicate the model has difficulty in capturing data profiles which are less common in the dataset.

\subsection{Unlawful social welfare Experiment}
\label{socialwelfare}
In the second experiment, the application of detecting risk profiles within a group of receivers of social welfare is considered.
People who have low income are qualified to claim these social assistance benefits from the social security system.
A part of the receiving group has no right to it, for example because they applied while being unaware they had no right to, or because they consciously commit fraud. 
This experiment investigates the  possibility of creating a binary classification model for detecting risk profiles in social welfare, in which the results are (path specific) counterfactually fair with respect to ethnicity.
This context is a central topic in the algorithmic fairness debate in the Netherlands.\footnote{https://www.volkskrant.nl/nieuws-achtergrond/een-druk-op-de-knop-van-de-computer-en-je-wordt-opeens-verdacht-van-fraude\textasciitilde b539dfde/}

\subsubsection{Data}
The analysis is conducted on 94.274 real data profiles of people receiving social welfare or being convicted for unlawfully receiving social welfare.
The source of the data are different governmental institutions in the Netherlands.\footnote{The data is provided strictly for scientific research purposes by Statistics Netherlands (CBS), and cannot be made publicly available. Reproducing of the results requires special license.}
The sample is taken from registered receivers and convictions in the Netherlands and balanced in the amount of convicted and non-convicted people. 
The sensitive variable is a binary ethnicity variable, indicating a western or non-western background based on the country of birth of the person and the country of birth of the parents of the person.
The labels, depicted as Y, are based on the financial claims sent by municipalities after conviction of unlawfully receiving social welfare.
The created data profiles consist of a wide selection of background variables, including gender, age, various income variables, education level and household information.
A full description of the variables and preprocessing is provided in the appendix.

\subsubsection{FairTrade implementation}
Domain knowledge to structure the causal graph is gained through literature, process analysis, and conversations with various domain experts. 
This includes social domain researchers, experts working in data processing for social welfare applications, and law enforcers in this context.
The structure in Figure \ref{fig:fairtrade_graph}c is well suited for the social welfare scenario. 
Ethnicity is a root variable which, according to literature, might (indirectly) influence various factors in the data profile \cite{tonry1997ethnicity, schwartz1976migration}.
A full assignment overview is provided in the appendix.\footnote{The assignments do not opinions or ideas of involved parties}
Interviews indicated that identification fraud\footnote{https://www.volkskrant.nl/nieuws-achtergrond/tot-vier-jaar-cel-voor-bulgarenfraude~b580f277/} could lead to discussion around resolving variables in this context.
The analysis leads to an expert based structure with little risk of excluding existing effects caused by the sensitive attribute. 
Furthermore, the setting is suitable for obtaining a sufficiently good representation of the unobserved confounders.
Combining information from municipalities, police registrations, tax registrations, debt registrations and other sources for a large sample of a country's population is an exceptional rich setting of information for a real data experiment.
A full description of the architecture, implementation and optimisation is provided in the appendix.

\begin{table}
\parbox{.35\linewidth}{
\centering
\caption{Classification accuracies for baseline models, with standard deviation over 20 repetitions with different random splits of 85.274 training points and 9.000 test points.}
    \begin{center}
    \begin{small}
    \begin{sc}
    \begin{tabular}{lcr}
    \toprule
    Baseline & accuracy \\
    \midrule
    MLP    & 0.691$\pm$ 0.007\\
    RF           & 0.683$\pm$ 0.003\\
    LR    & 0.690$\pm$ 0.004 \\
    \bottomrule
    \end{tabular}
    \label{tab: baselines}
    \end{sc}
    \end{small}
    \end{center}
}
\hspace{15pt}
\parbox{.48\linewidth}{
\caption{Classification accuracies and statistical parity scores for the auxiliary network under different fairness constraints }
\begin{center}
\begin{small}
\begin{sc}
\begin{tabular}{lccr}
\toprule
Input Aux. & Accuracy & Stat Par score  \\
model & & \\
\midrule
Z         & 0.547$\pm$ 0.005 & 0.981$\pm$ 0.008\\
Z,B       & 0.586$\pm$ 0.005 & 0.977$\pm$ 0.008\\
Z,B,R*     & 0.584$\pm$ 0.005 & 0.978$\pm$ 0.008\\
Z,B,R,X   & 0.668$\pm$ 0.006 & 0.964$\pm$ 0.011\\
Z,B,R,X,A & 0.668$\pm$ 0.006 & 0.949$\pm$ 0.012\\
\bottomrule
\end{tabular}
\label{tab: aux outcomes}
\end{sc}
\end{small}
\end{center}
}
\end{table}

\subsubsection{Results}
\begin{wrapfigure}[21]{r}{0.5\textwidth}
    \centering
    \includegraphics[scale=0.54]{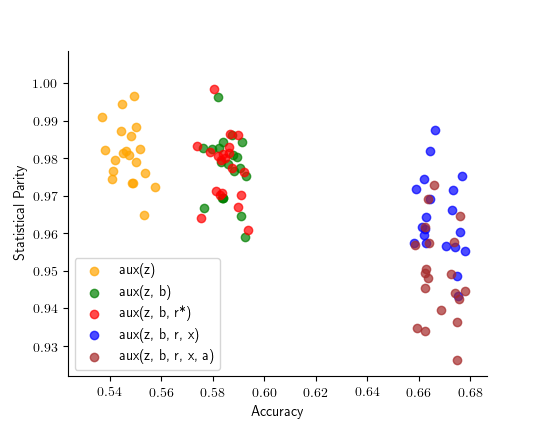}
    \caption{Statistical parity score and accuracy under different fairness constraints} 
    \label{fig:aux curve}
\end{wrapfigure}
An important assumption in the proposed CEVAE model is that the latent space becomes marginally independent of the sensitive attribute.
Visual inspection of TSNE scatter plots at different points during training suggest that the distinction between the latent spaces successfully disappears during training.
In order to put the prediction results in perspective, Table \ref{tab: baselines} provides the accuracy of a Multi-Layer Perceptron (MLP), Random Forest (RF), and a Logistic Regression (LR). 
Stable values between the accuracy of 0.68 and 0.70 are observed, indicating what accuracy can be expected without fairness constraints on the balanced dataset.
Next to accuracy, resulting prediction models are evaluated on fairness.
Ideally this would be done by directly evaluating if unfair path specific effects influence the prediction model outcomes.
However, in this case it is practically impossible to obtain the required data for this.
The `what if' scenario for intervening on someone's ethnicity is not observable.
In order to partially evaluate the fairness, a statistical parity based metric is used: $Statistical Parity Score = 1 - | \E[round(\hat{Y})=1|a=0]-\E[round(\hat{Y})=1|a=1]|$ which returns 1 in the case of perfect statistical parity.
Note that if a fair path exists from the sensitive variable to the outcome, a fair model is not expected to score 1 on this metric. 
The prediction model is compared for 5 versions of different constraints, in which the following input selections are considered: $[(Z), (Z,B), (Z,B,R^*), (Z,B,R,X), (Z,B,R,X,A)]$, 
with $R^* = R(Z,B,A,X(Z,B,a'))$.
The auxiliary model with $(Z,B,R^*)$ as input represents the situation in which none of the causal paths from $A$ to $Y$ is allowed, except for $A \rightarrow R \rightarrow Y$.
Hence, the value of $R$ should not obtain sensitive information through $X$, explaining the intervention which takes out the effect from $A$ to $X$.
The results are shown in Table \ref{tab: aux outcomes}, and visualised in Figure \ref{fig:aux curve}.
The model with only $Z$ as input achieves an accuracy of 0.55, indicating that $Z$ stores some but limited useful information about the covariates.
The models with input $(Z)$ and $(Z,B)$ satisfy the theoretical restriction of counterfactual fairness, excluding all paths from $A$.
From the results, one can clearly see the positive effect of implying fairness constraints on the group equality, although it comes at the cost of some accuracy performance.

\section{Black box auditing}
\label{black-box-scoring}
The FairTrade method focuses on creating models in line with fairness constraints.
Apart from making new fair models, there is interest in evaluating the fairness of existing models.
For example, recent calls in Dutch politics argue for a supervising body for all existing government algorithms.\footnote{https://nos.nl/artikel/2289495-d66-en-cda-willen-richtlijn-en-toezichthouder-voor-overheidsalgoritmes.html} 
We are not aware of any present method which could perform a satisfactory fairness audit.

We propose the following in order to adress these demands.
Consider a black box model which uses data profiles as input.
First, we need an estimate of the causal mechanisms in the data, for which we train a CEVAE model using the first 2 steps of the FairTrade method.
Hence, the CEVAE model says something about the data which goes into the black box model.
We use the CEVAE model to generate two datasets which differ only on aspects which the black box model should be invariant to.
If the black box model adheres to the posed constraints, predictions for the two datasets should approximately be the same.
Hence, the difference between the two sets of predictions can provide us with a fairness evaluation score.
This simple idea leverages the capabilities of the FairTrade method in a way that allows to audit black box (non-causal) models, as long as the generated datasets are sufficiently accurate.
In the appendix we include explanation on the practical execution of this audit, a simple simulation experiment to illustrate the idea, and consider related literature.

\section{Discussion}
\label{disucssion}

In this work, the FairTrade method is proposed to improve the fairness of ML predictions in practical scenarios.
The results indicate increased control over causal relations, and show the increased equality in treatment between sensitive groups when more fairness constraints are added.
We use an approximate method, bound to assumptions, in order to improve on existing solutions.
Requiring 100\% correctness of predictions and assumptions is not realistic for any real application of causal inference in the social domain. 
Especially under such circumstances, justifying the choices and pointing out corresponding imperfections is essential to work towards more fair methods.

First, although the aim of creating fairness in ML models is becoming more tangible with the arrival of causal based metrics and path specific considerations, there is no consensus on what a fair model is. 
Correcting statistics to prevent unequal outcomes is mainly deemed justified if the inequality is due to observational bias, but the justification becomes less clear in the case of population level inequality, or a mixture of the two. 
This challenge remains case specific, and needs further consideration on both a technical and an ethical level. 
The FairTrade method shows capable of improving fairness in practical applications, but is bound to some limitations throughout the steps.
In the first step, the causal assumptions can only capture a simplified structure of reality which is hard to verify.
In the second step, doing inference using a VAE setup is an approximation, with limited guarantees on recovering the true effects. 
Finally, in the last step, the causal based fairness of a predictive model can seldom be completely evaluated due to the impossibility of obtaining counterfactual observations.
More elaborate sensitivity analysis, and expansion of investigated practical case studies would be valuable future research.

\section*{Acknowledgments and Disclosure of Funding}
This research started as a master thesis at the University of Amsterdam.
After a period of theoretical research, a collaboration with the municipality of Amsterdam and Statistics Netherlands (CBS) was started in order to study a relevant problem with real data.
After finishing the thesis\footnote{https://scripties.uba.uva.nl/scriptie/695411}, the research continued as part of a project on \textit{fair algorithms}.
This project was carried out in a collaboration between CBS, municipality of Amsterdam, the University of Amsterdam, codefor.nl, the Association of Netherlands Municipalities (VNG) and other Dutch municipalities and was funded by the Ministry of the Interior and Kingdom Relations (BZK).

Special thanks to Tamas Erkelens, Barteld Braaksma and Gerhard Dekker for making the collaboration possible, and their enthusiastic involvement along the way.
We thank Joris Mooij as chair of the defence committee of the original master thesis.
Thanks to Anna Mitriaieva for the insightful discussions and suggestions.
Many thanks to Leon Willenborg, Koen Helwegen and Laurens Samson for proofreading earlier versions of the paper.
\newpage

\bibliography{bibliography}
\bibliographystyle{plainnat}

\newpage
\section*{Appendix}
All code, with the used hyperparameters as default arguments, can be found here:\\ \url{https://github.com/rik-helwegen/FairTrade}

\subsection*{Data generative process outcomes IHDP}
For the data generative process of the outcomes (Y) and treatmens (T) we follow the lines of \citet{madras2019fairness}, provided in algorithm 2 and 3. The used code and parameter values are provided in the code base under \path{ihdp_experiment/generateIHDP.py}.

\subsection*{Social Welfare experiment information}
Information for this experiment, as provided below, is also included in a technical report by the authors.\footnote{https://scripties.uba.uva.nl/scriptie/695411}

\subsubsection*{Definition sensitive variable (Ethnicity)}
The ethnicity variable in this experiment, denoted as $A$, is simplified to a binary variable, indicating a western or a non-western country of origin of the individual. 
This does not capture the complete meaning of ethnicity, but creates clarity in the analysis.
The value is based on the country of birth of a person and the country of birth of his or her parents.
For a first generation immigrant, his or her country of birth is leading. For a second generation immigrant, the country of birth of the mother is leading, except if this is the Netherlands, in which case the country of birth of the father is leading.
Due to border and name adjustments through time, the country of origin can differ from what it would be under current naming conventions. 
The according division between western and non-western countries is as follows:
\begin{enumerate}
    \item \textit{Non-western countries}: countries in Africa, Latin-America, Asia (excluding Indonesia and Japan), Morocco, Turkey, Suriname, Dutch Antilles, Aruba
    \item \textit{Western countries}: countries in Europe (excluding Turkey), North-America, Oceania, Indonesia and Japan
\end{enumerate}

\subsubsection*{Definition label (Unlawful receiver of social welfare)}
The labels are the basis on which the model learns how to recognise lawful and unlawful social welfare receivers.
The accuracy of the final model reflects how often the risk profile classification equals the actual labels.
The actual labels are based on financial claims send by the municipalities. 
A person obtains the label value 1 when he or she \textit{received a claim to repay unlawful social welfare}. 
For all other individuals, the label value is set to 0.
Hence, this is a setting with only positive and unknown labels, for which all unknowns are assumed to be negative.
The downside of this approach is that the negative labelled group will in fact include individuals which should have a positive label.
Information leakage of the label into the covariates is a considerable risk, as a conviction will cause the person to loose his or her social welfare, hence resulting in a change in income and possibly other attributes.
In order to mitigate this, measurements of the covariates from before January 2015 are used, and all positive labels are based on convictions which have been ruled after January 2015.

One consideration with the defined labels is leakage of label information into the covariates.
After a person is found guilty of unlawfully receiving of social welfare, the payments of the social welfare will stop.
This action thus has an effect on the income of this person, which decreases by the omitted amount of social welfare. 
In the modelled scenario, a directed effects towards the label are assumed, and none come from the label.
Therefore, this leakage of information would be in contradiction with the model assumptions.
In order to prevent this problem, all covariates are measured before January 2015 and all label information is measured related to the period after January 2015.
In specific, only positive labels are considered for which the social welfare is decided to be unlawful after Januray 2015. 
In many cases, the claim to return the social welfare followed months or even years after.
\\\\
\subsubsection*{Variables and preprocessing}
The data profile of individuals consists of the following list of variables with the according preprocessing. 
Prior to each variable or group of variabels, a letter B, R or X is provided. 
These letters refer respectively to \textit{Base} variables, \textit{Resolving} variables and \textit{other} variables, and indicate the position of the variables in final model structure.
The variable \textit{age} was initially thought of as base variable, but is, due to the lack of marginal independence with $A$, modelled as covariate instead.
\begin{enumerate}
    \item [B] \textit{Gender of partner}: consisting of two binary variables for having a male or a female partner. If no partner is registered, both variables obtain the value 0.
    \item [B] \textit{Mutation of purchasing power}: estimate of the change of purchasing power over the past year. 
    Unknown values are set to 0, indicating no measured change in purchasing power, this affected 0.1\% of the data records. The variable has a long tailed distribution, and is capped at 100\% difference in purchasing power, the variable is subsequently scaled between [-1,1] by dividing by this maximum value.
    \item [X] \textit{Age}: the age of the person, grouped in 10 year categories. The categories are included as a binary representation, in line with a one-hot encoding.
    \item [X] \textit{Education level}: indication of someone's highest level of achieved education.
    The grouping distinguishes 18 levels of education, which again is represented as a one-hot encoding. It is possible to have none of the education levels activated, i.e. set to 1.
    \item [X] \textit{Personal Primary Income}: the quantity of the main source of the income for the person.
    Unknown values are set to 0, which affects about 0.1\% of the data. 
    To standardise income, the income is capped between on the interval [-100k, 100k], and subsequently divided by 100k.
    Capping the income prevents a collapse of the standardised values due to outliers.
    As the group of interest is in a lower income regime, with an average of 1.5k, variation at this level is important to maintain.
    \item [X] \textit{Income Percentile before taxes}: the percentile group of the person relative to the Dutch population, measured before tax. The percentile assignments are by definition uniform over the population, the main mass of the distribution for people receiving social welfare is between the 10th and 40th percentile. The variable is grouped into the following categories:
    \begin{enumerate}
        \item INPP\_other: this person lives in household without observed percentile, has no personal income, or lives in institutional household
        \item INPP\_0\_20: this person falls within the percentile range $[0,20]$
        \item INPP\_20\_30: this person falls within the percentile range $(20,30]$
        \item INPP\_30\_100: this person falls within the percentile range $(30, 100]$
    \end{enumerate}
    Profiles contain a binary indication for each of the categories, creating a one-hot encoding representation.
    \item [X] \textit{Income from work abroad}, capacity of financial income from working in a foreign country. The variable is capped at 100k in order to prevent a collapse of the smaller amounts after normalising. 
    \item [X] \textit{Income from social benefits abroad}, capacity of financial income from social benefits in a foreign country. The variable is capped at 100k in order to prevent a collapse of the smaller amounts after normalising. 
    \item [X] \textit{Other income variables}: a selection of 10 other income metrics. The metrics include income before taxes, self-employment deduction, tax-free income, tax-exemption, striking deduction and profit margin deduction. 
    For these variables unknown values are set to zero, a natural baseline for financial quantities. The variables are divided by the absolute maximum in order to rescale the values to the interval $[-1,1]$.
    \item [X] \textit{Foreign parents}: the number of parents with a foreign country of origin. Three binary variables encode the possible values of 0, 1 and 2. 
    \item [X] \textit{Other social benefits}: binary indication for 9 social benefits on whether this person receives the benefit or not. Unknown values are put to zero. All benefits are provided by governmental institutions. The benefits include total incapacity income provision, youth handicapped provision, incapacity insurance, incapacity insurance self-employed, social welfare, partial incapacity provision and its special cases for elderly and self-employed people, and finally unemployment provision.
    \item [X] \textit{Gender partner}: represents the gender of a registered partner. This is encoded as two binary variables indicating the presence of a male partner and a female partner. If a person does not have a partner neither of the variables is activated.  
    \item [X] \textit{Household type}: indicating the type of household the person lives in. Binary variables for 8 household types are included. The possible household type ares: One person household, non married couple without children, married couple without children, non married couple with children, married couple with children, one parent household, other, institutional household. 
    \item [X] \textit{Property value}: the registered value of owned property. 
    If the person owns a house, the WOZ registration value of this property is given. 
    The value is capped at 500K, and standardised by dividing by 500k. 
    This is done to maintain variation after standardisation for the lower property valuations, as the main target group of the model has property in the lower range of the housing price spectrum.
    \item [R] \textit{Partner with debt}: binary indication of having a partner with debt. If a partner is under under supervision for having a too large debt, this is indicated with a binary variable.
    \item [R] \textit{Crime involvement}: indicates past involvement in different kind of crimes. Three binary variables indicate involvement in weapon related, drug related and other types of crime. Multiple of the variables can be activated for a single person. 
    \item [R] \textit{Recidivism}: indicating repeating involvement in crimes. A binary variable indicates if the person is labelled as recidivist. 
\end{enumerate}
All variable values are concatenated to create an individual's profile.
This way, a profile becomes a vector with 78 value entries, in which each category of the one-hot encodings are counted as a single entry. All model and variable assignment choices are made for research purposes. The model is not used in practice, and the choices do not reflect real ideas or opinions of any party. 

\subsubsection*{Architecture and implementation details}
For the implementation of the method we follow the line of \citet{louizos2017causal}.
The Z space is defined as a product distributions of Gaussians with dimensionality $D_z$.
The network has a single hidden layer of 100 nodes with ELU activation \cite{clevert2015fast}. The mean and variance have separate output nodes, in which the variance is ensured to be positive using a softplus activation.
The generative network reconstructs the data profiles, and thus provides the parameters for the reconstruction distributions.
Binary variables are modelled as bernoulli distributions, continuous variables as normal distributions and categorical variables as categorical distributions.
Furthermore, the generative networks are structured as TAR networks, separating output heads for the values of the sensitive attribute \cite{shalit2017estimating}.
The functions $p(\boldsymbol{x}|\boldsymbol{z},\boldsymbol{b},a)$, $p(\boldsymbol{r}|\boldsymbol{x},\boldsymbol{z},\boldsymbol{b},a)$ and $p(y|\boldsymbol{r},\boldsymbol{x},\boldsymbol{z},\boldsymbol{b},a)$ are parameterised as separate neural networks, each with an hidden layer of 100 nodes with ELU activation. The standard deviations of Normal distributions are truncated at a minimum of 0.1 for stability purposes. 
The joint model is optimised using ADAM \cite{kingma2014adam} with a learning rate of 1e-4, batch size of 512, and a single sample of $Z$ to approximate the expectation in the lowerbound.
The CEVAE optimisation takes around 50 minutes per repetition on a hyper threaded quad core machine.

For the final step of the FairTrade method, creating Fair predictions, an auxiliary neural network with one hidden layer of 100 nodes and ReLu activation is used \cite{nair2010rectified}. The output layer has a sigmoid activation function. 
The objective function is defined as the binary cross-entropy loss of the predicted Y values, and the accuracy is based on the number of correct predictions when the outcomes are rounded to binary values.
The model is optimised using RMSprop \cite{hinton2012lecture}.

\begin{figure}[H]
        \centering
        \includegraphics[width=0.45\textwidth]{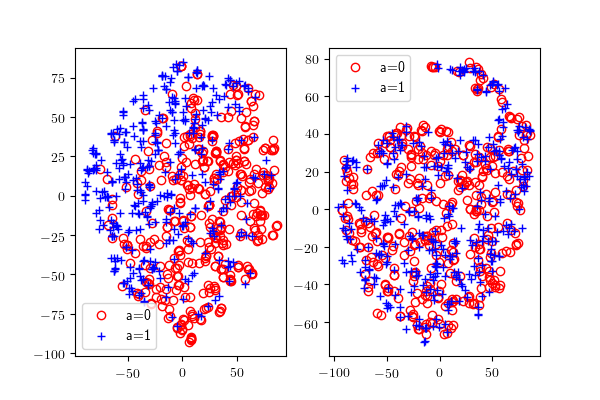}  
        \label{tsne}
        \caption{TSNE of the $Z$ space conditioned on $A$ after 1 iteration (left) and after 1000 iterations (right). As the two conditional distributions become less distinguishable, this shows the emergence of independence between $A$ and $Z$.}
\end{figure}

The TSNE plots in Figure \ref{tsne} show the inferred Z space after 1 and 1500 iterations, conditioned on the sensitive attribute. As expected, the distinction of the conditioned latent spaces disappears during training.

\subsection*{Black box scoring experiment}
To generate the data for auditing, the first two steps of the FairTrade method are followed in order to construct and train a CEVAE model.
Subsequently, the inference step is performed with the observed data. 
Using the inferred latent space, two reconstructions are generated, one for the standard forward pass and one with a counterfactual forward pass, changing the sensitive attribute.
This last action can be set up such that the two datasets differ only in terms of the path specific effects which are marked unfair.

A simple simulation experiment is performed in order to illustrate the evaluation of black box models using the estimate of causal mechanisms.
The experiment departs from the causal graph as shown in Figure \ref{fig:fairtrade_graph} (a). 
The sensitive variable is binary, the latent confounder is standard normal, $X$ consists of three normal distributions with different parameters and the outcome is binary. 
The code for the generating process and experiments are included in the code base.
The following data generate process is used in order to generate data:
\begin{align}
\begin{gathered}
    A \sim Ber(p_a)\\
    Z \sim N(\mu_z,\sigma_z)\\
    x_{1,i} \sim N(-(\gamma_x + a_i), \max(a_x, b_x + c_x \cdot z_i))\\
    x_{2,i} \sim N(z_i, \max(a_x, b_x + c_x \cdot z_i))\\
    x_{3,i} \sim N(\gamma_x + a_i, \max(a_x, b_x + c_x \cdot z_i))\\
    p_y = \gamma_y + \theta_a \cdot a_i + \theta_x \cdot x_i^2 + \theta_z \cdot z_i\\
    y_i  \sim Ber(\sigma(p_y))
\end{gathered}
\end{align}
In which $\sigma()$ is the sigmoid function, and the following parameter values are used:
\begin{equation*}
\begin{aligned}[c]
    a_x = 0.1 \\
    b_x = 0.55 \\
    c_x = 0.2 \\
    p_a = 0.5 \\
    \mu_z = 0 \\
    \sigma_z = 1 
\end{aligned}
\qquad\qquad
\begin{aligned}[c]
    \gamma_x = 1.5\\
    \gamma_y = -8.5\\
    \theta_a = 3 \\
    \theta_x = 2/3 \\
    \theta_z = 2  
\end{aligned}
\end{equation*}

The goal of this experiment is to test if 1) the CEVAE is cable of successful inference and reconstruction of the (unobserved) data, 2) the trained model can generate counterfactual distributions, and 3) black box models can be scored to compare the counterfactual fairness performance.

\begin{figure}[H]
    \centering
    \includegraphics[scale=0.6]{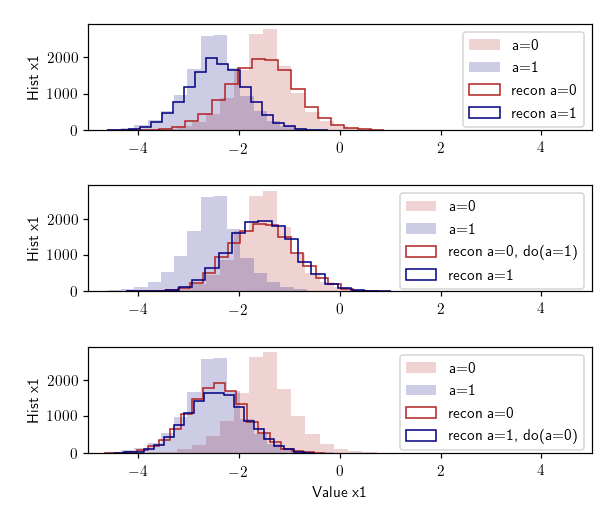}
    \caption{For one of the covariates in the simulation study the reconstruction distribution and two interventional distributions are shown, all conditioned on the sensitive variable. In the back the true values of x1 are shown.} 
    \label{fig:sim intervention}
\end{figure}

After the data is generated, $Z$ is dropped and a CEVAE model is fitted to the data. 
During training, the regularisation term goes to zero, and the log probabilities of X en Y increase and stabilise. 
The reconstruction distributions for $X_1$ are shown in in the top window of \ref{fig:sim intervention}.
The reconstructions are not perfect, but come close to the original data distributions.
For the second hypothesis we look at the interventional distributions, shown in Figure \ref{fig:sim intervention}. In accordance with the data generative process, intervention on $a$ determines the location of the distribution. 
The covariates $X_2$ and $X_3$ show similar performance.
For the last objective, three `black-box' models are considered, a logistic regression (LR), a second LR in which the predictive effect of the sensitive variable is set to a constant, and a Random Forest.
We use regression models in order to create an obvious difference in fairness by blocking the direct effect of $A$ in adjusted LR.
The counterfactual fairness score is expected to decrease as a result of this adjustment.
A test set is held apart from the beginning, the train set is used to train the CEVAE and to fit the black box models. 
The counterfactual fairness score is obtained by comparing the results of predictions based on the reconstructed test set, and the predictions of the counterfactual reconstructed test set. 
To create the counterfactual set, the value of the sensitive variable is switched for all individuals.
The outcomes are shown in Table \ref{tab: counterfact scores sim}. 

\begin{table}[H]
\caption{Counterfactual fairness scores for black-box models for a trained CEVAE, with STD for 20 repetitions including new samples from the CEVAE}
\vskip 0.15in
\begin{center}
\begin{small}
\begin{sc}
\begin{tabular}{lcr}
\toprule
Model & CF score \\
\midrule
Logistic Regression    & 0.71 $\pm$ 0.002\\
Logistic Regression - fixed a    & 0.45 $\pm$ 0.002\\
Random Forest   & 0.92 $\pm$0.028\\
\bottomrule
\end{tabular}
\label{tab: counterfact scores sim}
\end{sc}
\end{small}
\end{center}
\vskip -0.1in
\end{table}
In accordance with our expectation, the adjusted logistic regression obtains a lower counterfactual fairness score compared to the original logistic regression.

The evaluation method for black box models is also considered in the context of detecting unlawful social welfare. 
An experiment is set up using the same steps as in above simulation experiment. 
As a sanity check, the accuracy of the black box models in the reconstruction scenario is compared to the accuracy of the black box model in the real data scenario.
The latter, on which the model is trained, yields an accuracy of 0.67. However, when reconstructing the data, the prediction accuracy only reaches 0.57 for the logistic regression and 0.54 for the random forest. 
This difference is considered too significant to attach interpretation value to counterfactual scores of these models.
Before using the reconstructions for other purposes, further analysis of the reconstruction errors and uncertainty is required in order to understand the drop in the black box models.

\subsubsection*{Literature related to the black box audit}
A number of approaches have been proposed which use counterfactual datasets to analyse models, mainly focused on explainability. 
\citet{wachter2017counterfactual} and \citet{mothilal2019explaining} aim to generate a set of counterfactual input points which leads to a different outcome of the model, without explicitly modelling causal mechanisms among attributes. 
\citet{coston2019counterfactual} suggest using counterfactual data to evaluate models on counterfactual fairness, but do not account for unobserved confounders and focus on treatment based counterfactuals rather than path specific effects.

\end{document}